\pdfoutput=1
\documentclass[11pt]{article}

\usepackage[final]{acl}

\usepackage{times}
\usepackage{latexsym}
\usepackage{graphicx}
\usepackage{hyperref,multirow}
\usepackage{tabularx}
\usepackage{booktabs}

\usepackage[T1]{fontenc}
\usepackage[utf8]{inputenc}

\usepackage{microtype}

\usepackage{inconsolata}

\title{BookAsSumQA: An Evaluation Framework for Aspect-Based Book Summarization via Question Answering}

\author{
 \textbf{Ryuhei Miyazato\textsuperscript{1}},
 \textbf{Ting-Ruen Wei\textsuperscript{2}},
 \textbf{Xuyang Wu\textsuperscript{2}},
 \textbf{Hsin-Tai Wu\textsuperscript{3}},
 \textbf{Kei Harada\textsuperscript{1}},
\\
 \textsuperscript{1}The University of Electro-Communications,
\\
 \textsuperscript{2}Santa Clara University,
 \textsuperscript{3}DOCOMO Innovations, Inc.,
\\
 \small{
   \textbf{Correspondence:} \href{miyazato@uec.ac.jp}{miyazato@uec.ac.jp}, \href{harada@uec.ac.jp}{harada@uec.ac.jp}
 }
}

\begin{document}
\maketitle
\begin{abstract}
Aspect-based summarization aims to generate summaries that highlight specific aspects of a text, enabling more personalized and targeted summaries. However, its application to books remains unexplored due to the difficulty of constructing reference summaries for long text. To address this challenge, we propose BookAsSumQA, a QA-based evaluation framework for aspect-based book summarization. BookAsSumQA automatically constructs a narrative knowledge graph and synthesizes aspect-specific QA pairs to evaluate summaries based on their ability to answer these questions. Our experiments on BookAsSumQA revealed that while LLM-based approaches showed higher accuracy on shorter texts, RAG-based methods become more effective as document length increases, making them more efficient and practical for aspect-based book summarization\footnote{\url{https://github.com/ryuhei-miyazato/bookassumqa}}.
% While LLM-based approaches achieved the highest accuracy, our results indicated that RAG-based methods are more efficient and scalable for large-scale, multi-aspect summarization.
% Using the generated QA dataset, we evaluated aspect-based summaries produced by several summarization methods and found that RAG-based approaches are the most suitable in terms of both accuracy and efficiency.
\end{abstract}

\section{Introduction}
Automatic summarization condenses long texts into concise and informative representations, allowing readers to grasp key information efficiently. Book summarization applies this to novels, which are often lengthy and complex. The progress of automatic book summarization has been accelerated by the release of the BookSum dataset \citep{Kryscinski:22}, which contains novels paired with human-written summaries.
With the growing volume of books, there is increasing interest in aspect-based summarization (ABS), which produces summaries tailored to specific aspects, such as themes or genres. Although ABS helps readers quickly access desired information and has been more actively explored in domains such as reviews \citep{Xu:23} and lectures \citep{kolagar-zarcone-2024-humsum}, its application to books remains relatively understudied.
This is mainly because summarization research relies on manually created reference summaries, and building evaluation datasets for long documents is a labor-intensive and costly process. The longer the original document and the greater the number of aspects, the higher the human and financial costs become. \par
% \begin{figure}[t]
%   \includegraphics[width=\columnwidth]{image/example.png}
%   \caption{Examples of aspect-based summaries of novels. Aspect-based summarization is not merely a compression of key points, but generates a summary focused on content related to the target aspect.}
%   \label{fig:example}
% \end{figure}
\begin{figure}[t]
  \includegraphics[width=\columnwidth]{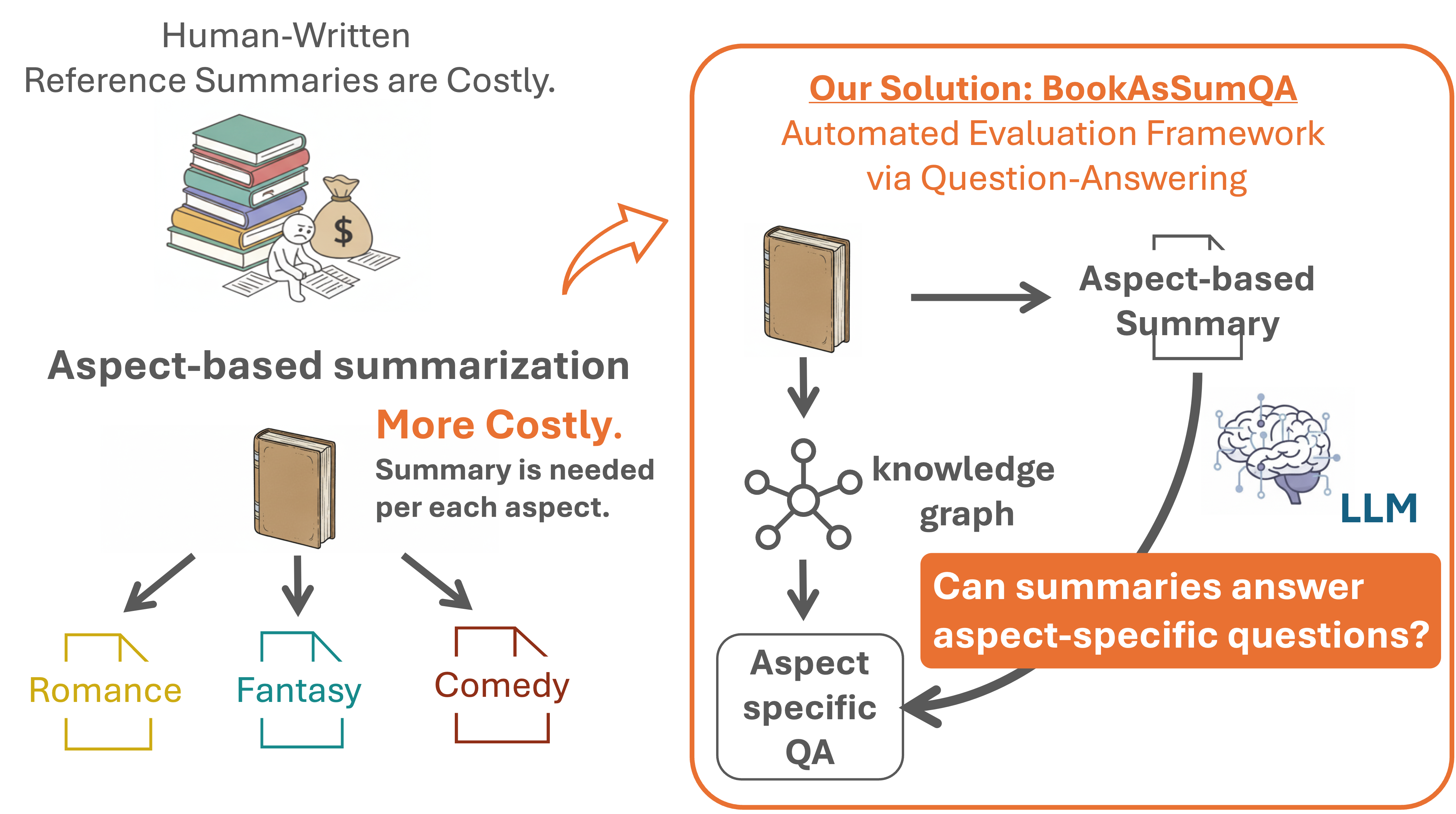}
  \caption{In BookAsSumQA, we generate aspect-specific QA pairs from a knowledge graph and evaluate summaries by testing whether they can answer these questions, thereby assessing aspect coverage without costly human-written references.}
  \label{fig:abs_example}
\end{figure}
To address this challenge, we propose BookAsSumQA, a QA-based evaluation framework for aspect-based book summarization that enables evaluation without manually created reference summaries. We synthesize aspect-specific QA pairs from the narrative through a knowledge graph, and evaluate aspect-based summaries by testing whether an LLM can answer these questions using the generated summary as reference. This allows us to measure how well the summary captures information about the aspects of the narrative. In this study, we define aspects as literary genres in novels (example: Figure~\ref{fig:abs_example}).  \par
First, we construct a knowledge graph that represents relationships among entities in the narrative. Using an LLM, we extract relationships between entities (e.g., characters) with a textual description, keywords, and an importance score, and incrementally upsert them into the graph to capture the global relationships within the narrative. 
Next, we construct aspect-specific QA pairs from the knowledge graph. To do so, we first identify edges that are relevant to a target aspect by calculating the cosine similarity between the text embeddings of the aspect term and the edge keywords, and then generate aspect-specific QA pairs based on the descriptions of those edges.
Finally, we evaluate ABS methods using the generated QA pairs by assessing whether each generated summary can correctly answer the questions. We then compare the generated answers against the ground-truth using ROUGE-1, METEOR, and BERTScore. By comparing the accuracy, we investigate which method is most suitable for aspect-based book summarization.

\section{Related Work}
In the field of book summarization, as the BookSum dataset \citep{Kryscinski:22} provides pairs of public domain novels and generic summaries, obtaining the summaries is well studied \citep{Wu:21, Xiong:23, Pu:23}. In this study, we focus on ABS, which generates summaries centered on specific aspects of a text. Unlike Query-Focused Summarization (QFS), which generates summaries in response to specific user queries (e.g., SQuALITY \citep{Wang:22}), ABS instead focuses on predefined aspects such as genres or themes. \par

ABS has been actively studied in domains such as news \citep{Zhang:22}, reviews \citep{Xu:23}, lecture materials \citep{kolagar-zarcone-2024-humsum}, and multi-domain documents \citep{hayashi-etal-2021-wikiasp}, where reference summaries are often manually created or readily available. However, for long documents like books, creating such references is labor-intensive and costly, limiting the application of ABS in this domain. \par

To overcome this difficulty, we propose a framework that evaluates aspect-based summaries of novels without manual reference summaries. 
While several studies have proposed reference-free evaluation metrics for summarization that assess summary quality without relying on gold reference summaries \cite{chen-etal-2021-training, liu-etal-2022-reference, gigant-etal-2024-mitigating}, we introduce a QA-based framework that evaluates summaries without manual references by generating QA pairs from the source text, measuring how much information from the source text is captured in the summary \cite{Hirano:01, scialom-etal-2019-answers, pu-etal-2024-summary}.
In this work, we further extend this approach by generating aspect-specific QA pairs to evaluate how well each aspect-based summary captures information related to its corresponding aspect in the original text.
% While previous studies have introduced QA-based evaluation methods for summarization (e.g., \cite{Hirano:01, scialom-etal-2019-answers, pu-etal-2024-summary}) to assess how much information from the source text is captured in the summary, our framework extends these methods by synthesizing aspect-specific QA pairs.

\section{BookAsSumQA}
% In this section, we describe the BookAsSumQA framework in detail.

\begin{figure}[h!]
  \includegraphics[width=\linewidth]{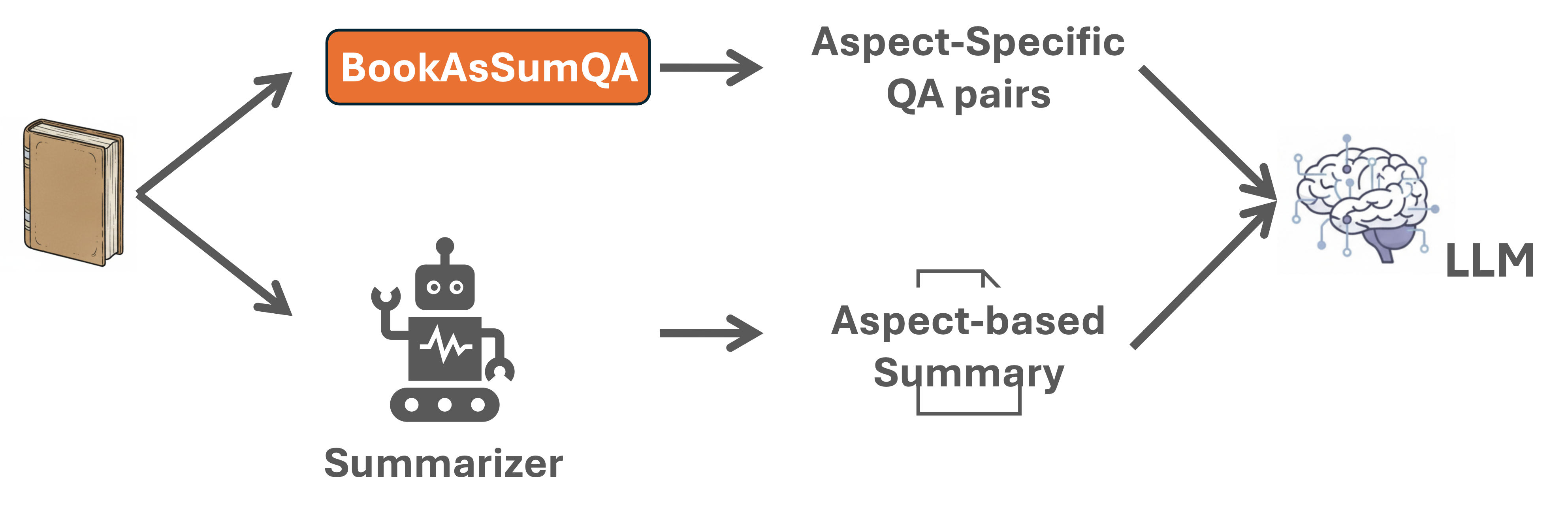}
  \caption{BookAsSumQA: Evaluation framework for aspect-based book summarization.}
  \label{fig:how_to_use_bookassumqa}
\end{figure}
\subsection{ABS Evaluation with BookAsSumQA}

% In BookAsSumQA, as illustrated in Figure~\ref{fig:how_to_use_bookassumqa}, we evaluate generated aspect-based summaries using QA. We first generate QA pairs related to the story, and then assess the quality of aspect-based summaries by measuring how accurately LLM can answer the aspect-specific QA pairs based on the generated summaries.
In BookAsSumQA (Figure~\ref{fig:how_to_use_bookassumqa}), we shift the evaluation of aspect-based summaries into a Question-Answering task. QA pairs are automatically synthesized through a knowledge graph of the narrative, where nodes are enriched with keywords and description to generate comprehensive aspect-specific questions. The quality of a summary is then assessed by measuring how well the generated aspect-based summary enables an LLM to answer these aspect-specific QA, indicating how much information about the target aspect the summary truly captures.

\subsection{QA Generation Process}
\begin{figure*}[t]
  \includegraphics[width=\linewidth]{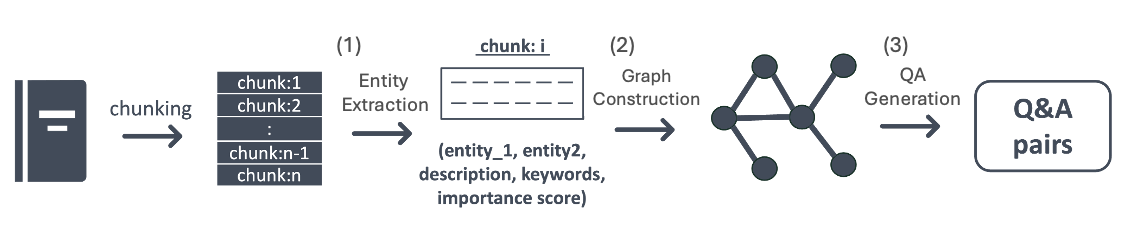}
  \caption{QA Generation Process. (1) splitting the text into chunks and extracting entities and relations, (2) inserting the extracted entities and relations into a knowledge graph as nodes and edges, and (3) synthesizing aspect-specific QA pairs from the completed graph. }
  \label{fig:example}
\end{figure*}

An overview of the QA generation process is illustrated in Figure~\ref{fig:example}. 
The process consists of three stages: (1) splitting the text into chunks and extracting entities and relations, (2) inserting the extracted entities and relations into a knowledge graph as nodes and edges, and (3) synthesizing aspect-specific QA pairs from the completed graph.   

\paragraph{(1). Chunking and Extraction}
Each book is split into chunks of 1,200 characters with an overlap of 100 characters, following the parameters of GraphRAG \citep{edge2024local}. From each chunk, entities (e.g., characters, events, concepts) are extracted using an LLM with a specifically designed prompt (2-shot, Appendix~\ref{sec:prompt}, Figure~\ref{fig:entity-extraction-prompt}). For each extracted relation, the prompt instructs the LLM to output a textual description, representative keywords, and an importance score ranging from 1 to 10, reflecting the importance of the relationship within the local context.

\paragraph{(2). Knowledge Graph Construction}
The extracted entities and relations are incrementally inserted into a knowledge graph, where each edge is labeled with keywords, a textual description, and an importance score. If an entity already exists, its information is updated and summarized as needed, with keywords regenerated accordingly. In addition, importance score is accumulated by adding the newly assigned value to reflect repeated or strengthened relationships across chunks.

\paragraph{(3). QA Generation}
 Once the knowledge graph is constructed, we generate aspect-specific QA pairs. We first filter edges to keep only those with an importance score of 10 or higher, considering relationships above this threshold to be important. An importance score of 10 indicates either a salient relationship or one that appears multiple times in the narrative, making it a stronger candidate for generating aspect-specific QA. From these, a maximum of 100 edges were selected. QA pairs are then generated from the description of each edge using a dedicated prompt (1-shot, Appendix~\ref{sec:prompt}, Figure~\ref{fig:qa-prompt}), with keywords from the edge also included in the generated QA. 
For each aspect, aspect-specific QA pairs were selected by calculating the cosine similarity between the text embeddings of the aspect and those of the QA keywords, and the top five most relevant QA were retained. Examples of aspect-specific QA pairs are also provided in Appendix~\ref{sec:example_of_generated_qa}.

We utilized GPT-4o-mini \footnote{\url{https://openai.com/index/gpt-4o-mini-advancing-cost-efficient-intelligence/}} for both entity extraction and QA generation and used sentence-transformers/paraphrase-MiniLM-L6-v2 ~\cite{reimers-2019-sentence-bert} for text embedding. For implementation of graph-generation, we referred to the code of LightRAG \citep{guo2024lightrag}.

\section{Experimental Settings}
\label{sec:main_experiment}

\subsection{Models}
Since no existing ABS method specifically targets books, we compare various approaches, including LLMs and RAGs. Detail information about the models is in Appendix~\ref{sec:summarizer_detail}.

\paragraph{LLMs}

Following the strategy of BooookScore \cite{chang2024booookscore}, we adopt two workflows for summarizing book-length documents that exceed the model’s context window: (1) Hierarchical Merging (Hier), which recursively merges summaries of individual chunks into higher-level summaries, and (2) Incremental Updating (Inc), which incrementally updates a single global summary as each new chunk is processed.
Detailed descriptions are provided in Appendix~\ref{sec:summarizer_detail}. \par

For experiments, we use both an open-source model, meta-llama/Llama-3.1-8B-Instruct \footnote{\url{https://huggingface.co/meta-llama/Llama-3.1-8B-Instruct}}, and a closed-source model, GPT-4o-mini.

\paragraph{RAGs}
RAG retrieves information relevant to a query from external sources and generates an answer. In this study, we adopt NaiveRAG \citep{Gao:23}, as well as GraphRAG ~\citep{edge2024local} and LightRAG~\citep{guo2024lightrag}, which employ graph structures to organize external information. 

\subsection{Setup}
The original texts used in this experiment are taken from BookSum~\citep{Kryscinski:22}, which sources books from the Project Gutenberg public domain book repository with expired copyrights.
We selected texts with varying lengths: over 200,000 words (large), between 90,000 and 110,000 words (middle), and less than 20,000 words (small), comprising 12, 9, and 9 books respectively, for a total of 30. 
\begin{table}[h]
\centering
\small
\begin{tabular}{|c|c|c|}
\hline
Fantasy        & Romance         & Comedy          \\  \hline
Paranormal     & Young Adult     & Horror          \\  \hline
History        & Action          & Science Fiction \\  \hline
Mystery        & Adventure       &  Crime              \\  \hline
Thriller       & Poetry          &                 \\  
\hline
\end{tabular}
\caption{List of Aspects used in this study.}
\label{tab:aspects}
\end{table}
In this paper, we define fourteen “aspects” as the literary genre of a novel with reference to Wikipedia's List of writing genres\footnote{\url{https://en.wikipedia.org/wiki/List\_of\_writing\_genres\#Fiction\_genres}}(see Table~\ref{tab:aspects}).\par

For each method, aspect-based summaries were generated for the aspects listed in Table~\ref{tab:aspects}, with each summary limited to 300 tokens. The generated summaries were evaluated based on their ability to answer the corresponding QA pairs with referring the generated summary. The prompts used for this QA-answering process are provided in the Appendix~\ref{sec:prompt} (Figure~\ref{fig:qa-answering}). The accuracy of the answers was evaluated using ROUGE-1~\citep{lin-2004-rouge}, METEOR~\citep{banerjee-lavie-2005-meteor} and BERTScore ~\citep{Zhang:20} metrics, measuring the alignment between the generated answers and the ground-truth.\par
RAG-based methods index the original text once and reuse it to generate summaries for different aspects, whereas LLM-based methods generate a new summary every time for each aspect.

\section{Results}
\subsection{Question Answering
\label{sec:qa_accuracy_using_aspect-based_summaries}
Using Aspect-Based Summaries}

\begin{table}[htbp]
\centering

\resizebox{\linewidth}{!}{%
\begin{tabular}{llccc}
\hline
\multicolumn{2}{l}{Type \quad method} & ROUGE-1   & METEOR  & BERTScore \\
\hline
\multirow{4}{*}{LLM} 
 & Llama + Hier  & 22.43  & 19.23  & 85.66      \\
 & GPT + Hier    & \textbf{22.49}  & \textbf{19.49}  & \textbf{85.82}     \\
 & Llama + Inc  & 21.91  & 18.23  & 85.48      \\
 & GPT + Inc    & 21.90  & 18.76  & 85.47      \\
\hline
\multirow{3}{*}{RAG}
 & NaiveRAG    & 21.43  & 18.66  & 85.44      \\
 & GraphRAG    & 14.66  & 13.56  & 84.50      \\
 & LightRAG    & 20.61  & 18.41  & 85.51      \\
\hline
\end{tabular}
}
\caption{Results of aspect-based summarization using different methods. LLM-based methods include Llama-3.1-8B-Instruct (Llama) and GPT-4o-mini (GPT).}
\label{tab:results1}
\end{table}

Table~\ref{tab:results1} shows the accuracy for aspect QA with generated aspect-based summaries. Each value represents the average result across all aspects. \par
Overall, the method that applies Hierarchical Merging with GPT-4o-mini achieved the highest scores. Among LLM-based methods, Hierarchical Merging was better than Incremental Updating, and LLM-based methods overall surpass RAG-based methods. For RAG, NaiveRAG achieves the best results, while GraphRAG shows considerably lower scores compared to the other methods.\par
One possible reason for the superior performance of LLM-based methods is that LLM-based methods extract aspect-specific information from finer-grained chunks. Although incremental updating incorporates previous context, using both the prior summary and the current chunk may make it harder to extract targeted information. In GraphRAG, summaries are generated for each community in the graph and used to answer QA, making it less effective at capturing aspect-related stories.
According to the results in the Appendix~\ref{sec:geneic_summary} (Table~\ref{tab:comparison_summary}), GraphRAG achieves the highest accuracy in conventional summarization, suggesting that improving the construction of the graph and the summarization process could lead to better scores in the future.

\subsection{Comparison by Original Text Length}
\begin{table}[h]
\centering
\resizebox{\linewidth}{!}{%
\begin{tabular}{llccc}
\hline
Size & Method & ROUGE-1 & METEOR & BERTScore \\
\hline
Small  & GPT + Hier   & \textbf{25.66} & \textbf{21.91} & \textbf{86.54} \\
       & GPT + Inc    & 24.81 & 20.84 & 86.14 \\
       & NaiveRAG    & 22.09 & 19.24 & 85.58 \\
\hline
Middle & GPT + Hier   & \textbf{21.95} & \textbf{19.52} & 85.56 \\
       & GPT + Inc    & 21.68 & 18.68 & 85.35 \\
       & NaiveRAG    & \textbf{21.95} & 19.45 & \textbf{85.62} \\
\hline
Large  & GPT + Hier   & 20.50 & \textbf{17.65} & \textbf{85.48} \\
       & GPT + Inc    & 19.88 & 17.27 & 85.06 \\
       & NaiveRAG    & \textbf{20.55} & 17.64 & 85.21 \\
\hline
\end{tabular}
}
\caption{Comparison by Original Text Length \\ (Small: <20k words, Middle: 90k–110k, Large: >200k)}
\label{tab:results2}
\end{table}

We conducted an experiment to compare summarization performance across different lengths of the original text. In this experiment, we used the best-performing models from the LLM-based and RAG-based approaches identified in Section~\ref{sec:qa_accuracy_using_aspect-based_summaries}. \par

As shown in Table~\ref{tab:results2}, the performance tends to decline as the length of the original text increases. Although NaiveRAG performs worse than the LLM-based method in the small group, its performance becomes comparable to that of the LLM-based approach in the middle and large groups. \par

Considering that RAG-based methods can generate aspect-based summaries for different queries with a single indexing of the original text, RAG-based approaches may be more suitable for aspect-based summarization of longer documents.

\section{Conclusion}
In this study, we proposed BookAsSumQA, a QA-based evaluation framework for aspect-based book summarization. Constructing knowledge graphs and automatically generating aspect-specific QA enable evaluation of ABS quality without human-annotated reference summaries.
In our experiments with BookAsSumQA, while LLM-based approaches performed better on shorter texts, RAG-based methods  achieved comparable performance on longer documents. These results suggest that RAG-based methods are more practical and scalable choice for aspect-based book summarization. Future work will explore specialized indexing and retrieval techniques.

\section*{Limitations}
This study has several limitations. First, we used gpt-4o-mini to generate QA pairs for summary evaluation; the choice of model may affect the evaluation results. In future work, we plan to investigate the impact of different models for QA generation. Second, both QA generation and answering relied on LLMs, which may incorporate external knowledge beyond the original text or summaries. To address this, we plan to explore methods for restricting the model’s context strictly to the given text and summaries, ensuring fairer evaluation.
Third, we have not yet compared our framework with other reference-free evaluation metrics or with human judgments. Such comparisons would help clarify how BookAsSumQA aligns with human evaluation and how it complements existing automatic metrics in terms of reliability and interpretability.

% In this study, we evaluate aspect-based summarization by generating QA pairs with ChatGPT to avoid the high cost of manually creating reference summaries for novels. By assigning aspects to the questions, we were able to evaluate how effectively the generated summaries capture information related to those aspects \par
% However, since the QA pairs are generated per chunk, they tend to be fragmented and do not capture the narrative’s coherence or the importance of key story elements. This means that the evaluation does not properly assess whether the summary captures important aspect-related information. Consequently, further work is needed to develop robust evaluation metrics for summarization quality and narrative understanding in long, complex texts.

\section*{Acknowledgments}
We would like to thank Dr. Shunsuke Kitada for his valuable advice and insightful feedback on the writing of this paper.

\bibliography{custom}

% \newpage

\appendix
% \onecolumn

\section{Experiment with Generic Summaries}

\subsection{Comparison Results between Reference Summaries and Standard Summaries}
\label{sec:geneic_summary}

\begin{table}[h!]
\centering
\resizebox{\linewidth}{!}{%
\begin{tabular}{llccc}
\hline
\multicolumn{2}{l}{Type \quad method} & ROUGE1 &  METEOR & BERTScore \\
\hline
\multirow{4}{*}{LLM}
 & GPT + Hier  & 20.64 &  9.87 & 82.89 \\
 & GPT + Inc    & 21.64 &  10.29 & 82.49 \\
 & Llama + Hier & 23.96 & 11.28 & \textbf{83.10} \\
 & Llama + Inc  & 24.03 &  11.21 & 82.60 \\
\hline
\multirow{3}{*}{RAG}
 & NaiveRAG    & 20.13 & 9.58 & 81.94 \\
 & GraphRAG    & \textbf{25.37}  & \textbf{14.78} & 80.29 \\
 & LightRAG    & 20.66 &  10.00 & 81.87 \\
\hline
\end{tabular}
}
\caption{Comparison Results between Reference Summaries and Standard Summaries.}
\label{tab:comparison_summary}
\end{table}

We conducted an experiment comparing the generic summaries generated by each model with the reference summaries in BookSum to evaluate the models’ capabilities for generic summarization. The results are shown in Table~\ref{tab:comparison_summary}.\par
In BookAsSumQA, the performance of GraphRAG was considerably worse than other methods. However, for standard summarization, it achieved the highest scores on two metrics based on character overlap. In contrast, it obtained the lowest score on BERTScore, which compares semantic similarity. \par

\subsection{Results of BookAsSumQA with Generic Summaries}
\label{sec:bookassumqa_geneic_summary}

\begin{table}[h!]
\centering
\resizebox{\linewidth}{!}{%
\begin{tabular}{llccc}
\hline
\multicolumn{2}{l}{Type \quad method} & ROUGE & METEOR & BERT\_Score \\
\hline
\multirow{4}{*}{LLM}
 & GPT + Hier  & 20.65 & 18.45 & 85.35 \\
 & GPT + Inc    & 20.63 & 17.51 & 85.23 \\
 & Llama + Hier & 19.86 & 16.45 & 85.23 \\
 & Llama + Inc  & 20.72 & 17.27 & 85.41 \\
\hline
\multirow{3}{*}{RAG}
 & NaiveRAG    & 19.76 & 17.28 & 85.05 \\
 & GraphRAG    & 15.12 & 14.37 & 84.81 \\
 & LightRAG   & 20.29 & 17.79 & 85.48 \\
\hline
\end{tabular}
}
\caption{The results of BookAsSumQA with generic summaries.}
\label{tab:bookassumqa_generic_summary}
\end{table}

We conducted an experiment comparing the accuracy of answering QA pairs generated by BookAsSumQA, using standard summaries produced by each model employed in our experiments in Section~\ref{sec:main_experiment}. The results are shown in Table~\ref{tab:bookassumqa_generic_summary}. \par
Compared to the results in Table~\ref{tab:results1}, aspect-based summaries achieved higher accuracy in answering aspect-specific QA. Additionally, while there were notable differences among methods when using aspect-based summaries, the results for generic summaries were more similar across methods. These findings indicate that BookAsSumQA serves as an evaluation framework for aspect-based summarization.

\section{Detail Information of Summarizer}
\label{sec:summarizer_detail}
\subsection*{LLMs}

\begin{figure}[h!]
  \includegraphics[width=\linewidth]{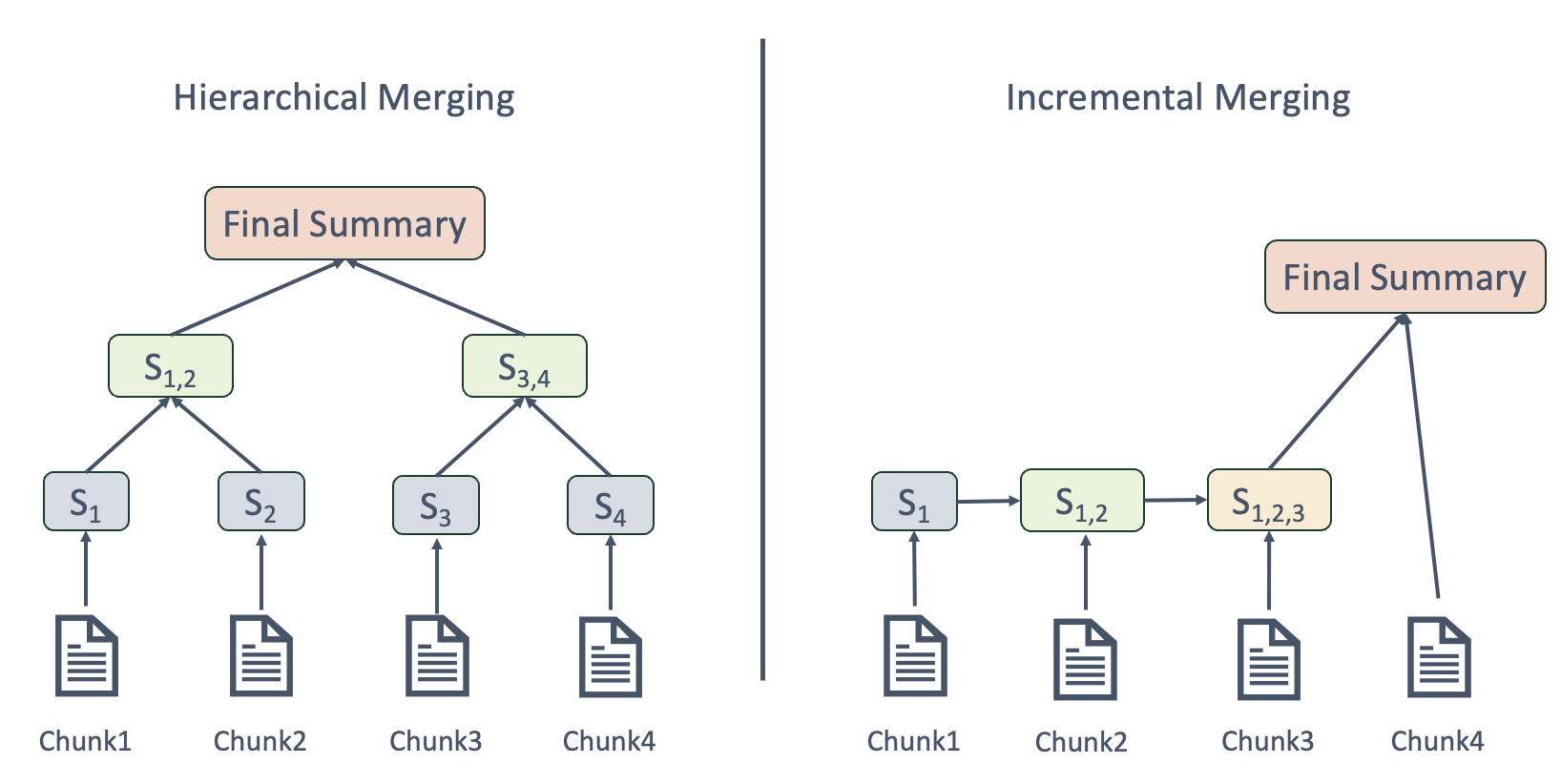}
  \caption{(1) Hierarchical Merging and (2) Incremental Updating.}
  \label{fig:llm-based-example}
\end{figure}

For LLM-based summarization, we adopt two prompting workflows for summarizing book-length documents that exceed the model’s context window (Figure~\ref{fig:llm-based-example}): (1) Hierarchical Merging (Hier) and (2) Incremental Updating (Inc), following BooookScore \cite{chang2024booookscore}.\par

In both workflows, the input document is first divided into smaller chunks (e.g., a chunk size of 2048 tokens). In the hierarchical merging strategy, each chunk is summarized separately, and the resulting summaries are merged using additional prompts. In the incremental updating strategy, a global summary is updated and compressed step-by-step as the model processes each chunk.\par

\subsection*{RAGs}
\begin{figure}[h!]
  \includegraphics[width=\linewidth]{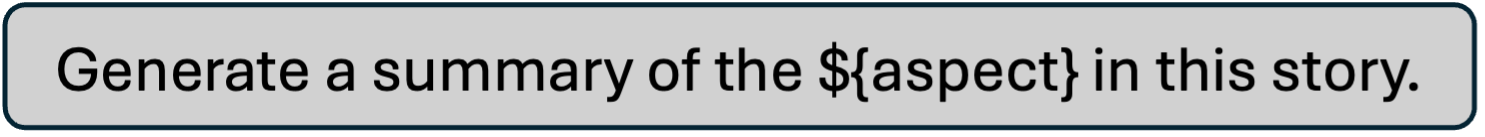}
  \caption{The query used for RAG-ased method.}
  \label{fig:rag_query}
\end{figure}

For RAG-based method, we used several RAG as described below. We used the default settings for indexing and retrieval methods, and built the same database for each aspect-based summarization approach. For each aspect, summaries were generated using query (Figure~\ref{fig:rag_query}) corresponding to that aspect as queries.

\begin{itemize}
\item \textbf{NaiveRAG \citep{Gao:23}} \\
NaiveRAG is a standard RAG system. It splits texts into chunks, embeds them, retrieves the most similar ones to a query, and generates an answer.

\item \textbf{GraphRAG \citep{edge2024local}} \\
GraphRAG creates a knowledge graph from the source text, generates community summaries by summarizing subgraphs, and uses them to answer queries.

\item \textbf{LightRAG \citep{guo2024lightrag}}\\
LightRAG builds a knowledge graph from the source text, retrieves relevant parts via the graph based on query keywords, and generates an answer.

\end{itemize}

\onecolumn

\section{Prompt}
\label{sec:prompt}

\begin{figure*}[h!]
  \includegraphics[width=\linewidth]{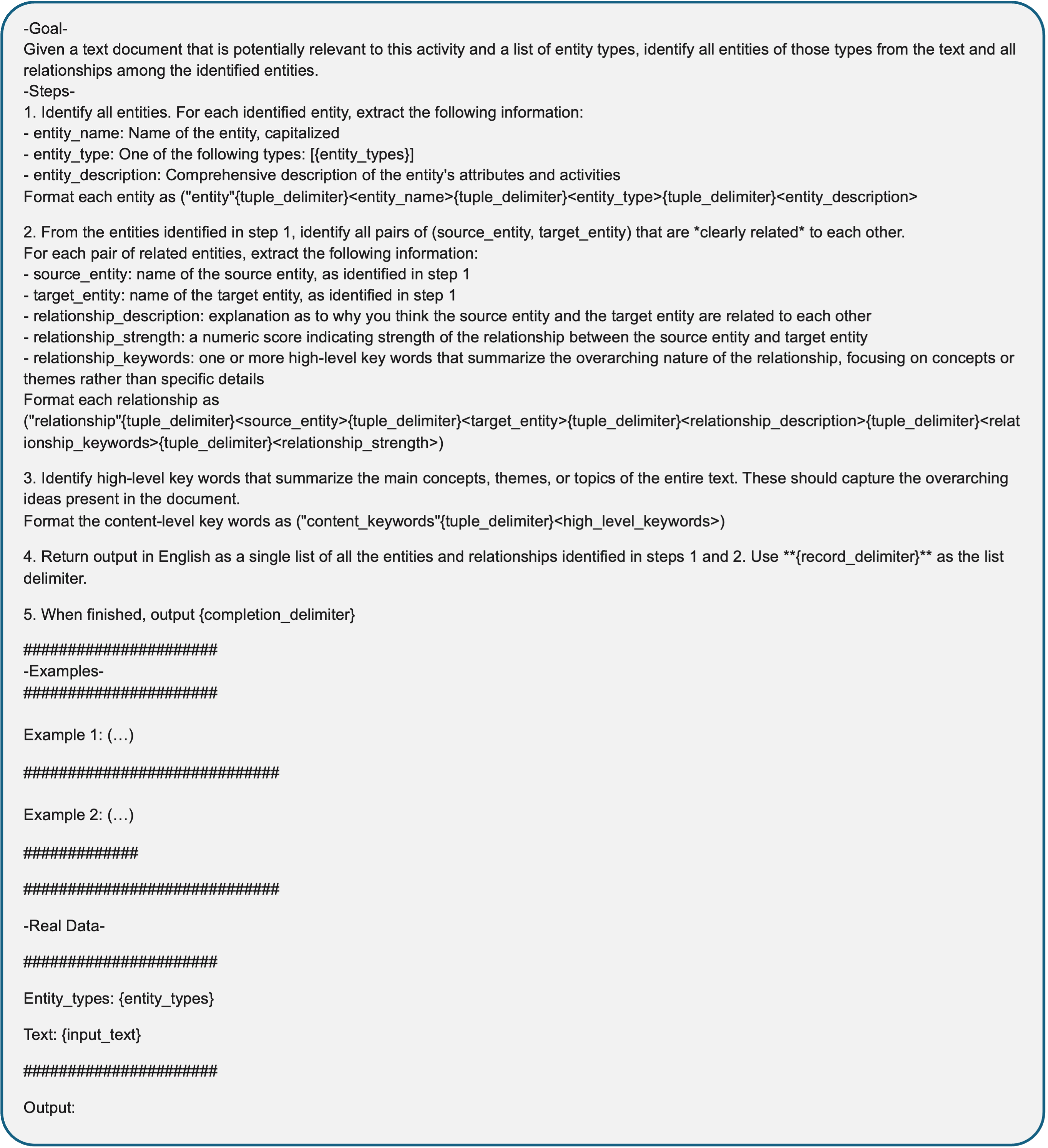}
  \caption{Entity extraction prompt (Vanity Fair).}
  \label{fig:entity-extraction-prompt}
\end{figure*}

\begin{figure*}[h!]
  \includegraphics[width=\linewidth]{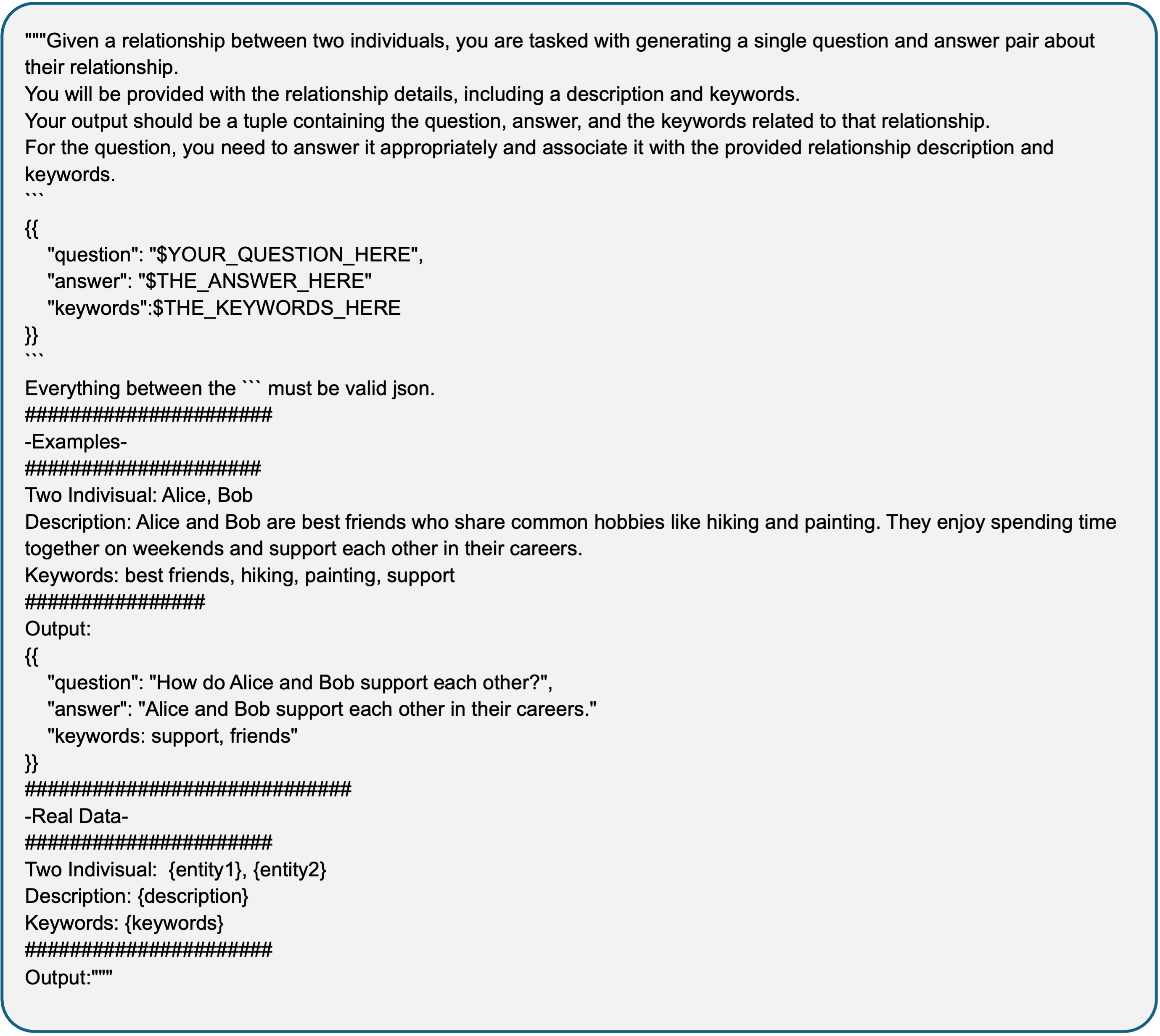}
  \caption{QA generation prompt (Vanity Fair).}
  \label{fig:qa-prompt}
\end{figure*}

\begin{figure*}[h!]
  \includegraphics[width=\linewidth]{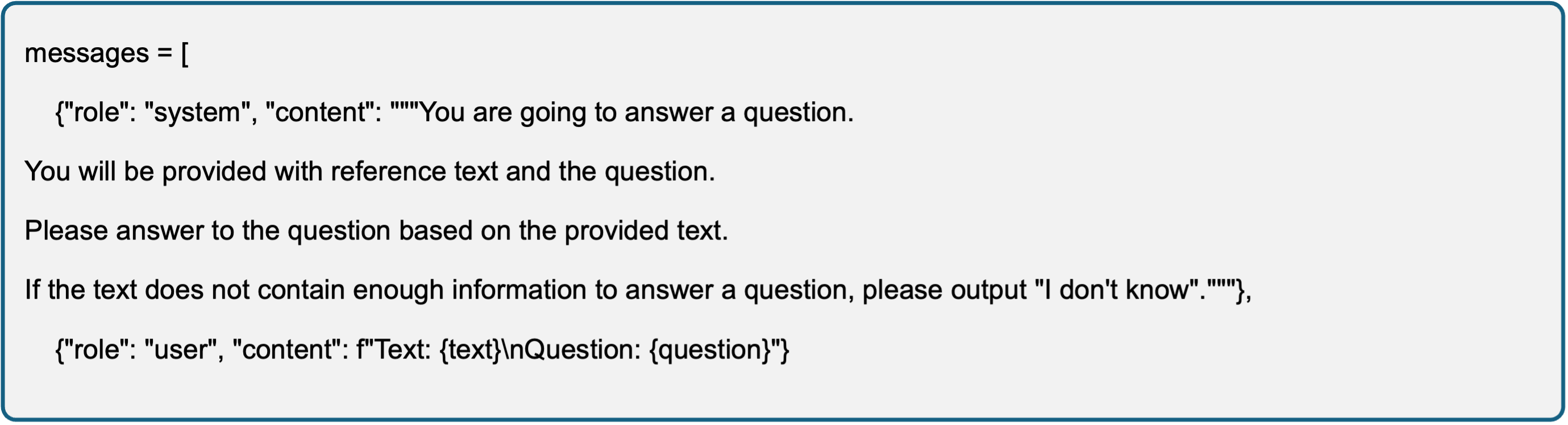}
  \caption{QA answering prompt (Vanity Fair).}
  \label{fig:qa-answering}
\end{figure*}

\clearpage
\section{Example of Generated QA}
\label{sec:example_of_generated_qa}
\begin{table*}[h!]
\centering

\begin{tabularx}{\textwidth}{l X X X}
\toprule
\textbf{Aspect} & \textbf{Question} & \textbf{Answer} & \textbf{Keywords} \\
\midrule
Romance & What indicates Joseph Sedley's romantic interest in Rebecca? 
& Joseph Sedley's romantic interest in Rebecca is indicated by his efforts to impress her, his sensitivity about his vanity, and his dependency on her during his illnesses, all of which suggest a growing intimacy and admiration for her.
& romantic interest, intimacy, admiration, dependency \\
\midrule
Action & What is the nature of the relationship between Miss Sharp and Miss Pinkerton? 
& The relationship between Miss Sharp and Miss Pinkerton is characterized by conflict and personal animosity, with Miss Sharp openly defying Miss Pinkerton's authority and expressing hatred towards her.
& conflict, authority, antagonism, rebellion, defiance \\
\midrule
Fantasy & What complexities characterize the friendship between George and Rawdon?
& Their friendship is characterized by playful interactions, rivalry in romantic interests, elements of manipulation, and shared gambling habits, which create both camaraderie and challenges.
& friendship, rivalry, manipulation, gambling, camaraderie \\
\midrule
Young Adult & What is the nature of the relationship between Rebecca and Lord Steyne?
& The relationship between Rebecca and Lord Steyne is multifaceted, characterized by mentorship, ambition, and social dynamics, with Rebecca leveraging Lord Steyne's favoritism for her family's benefit while also being dependent on his financial support.
& mentor-mentee, social dynamics, influence, ambition \\
\bottomrule
\end{tabularx}
\caption{Aspect-based QA Examples (Vanity Fair).}

\end{table*}

\clearpage
\section{The Example of Aspect-Based Summary (Romance)}

\begin{figure*}[h!]
  \includegraphics[width=\linewidth]{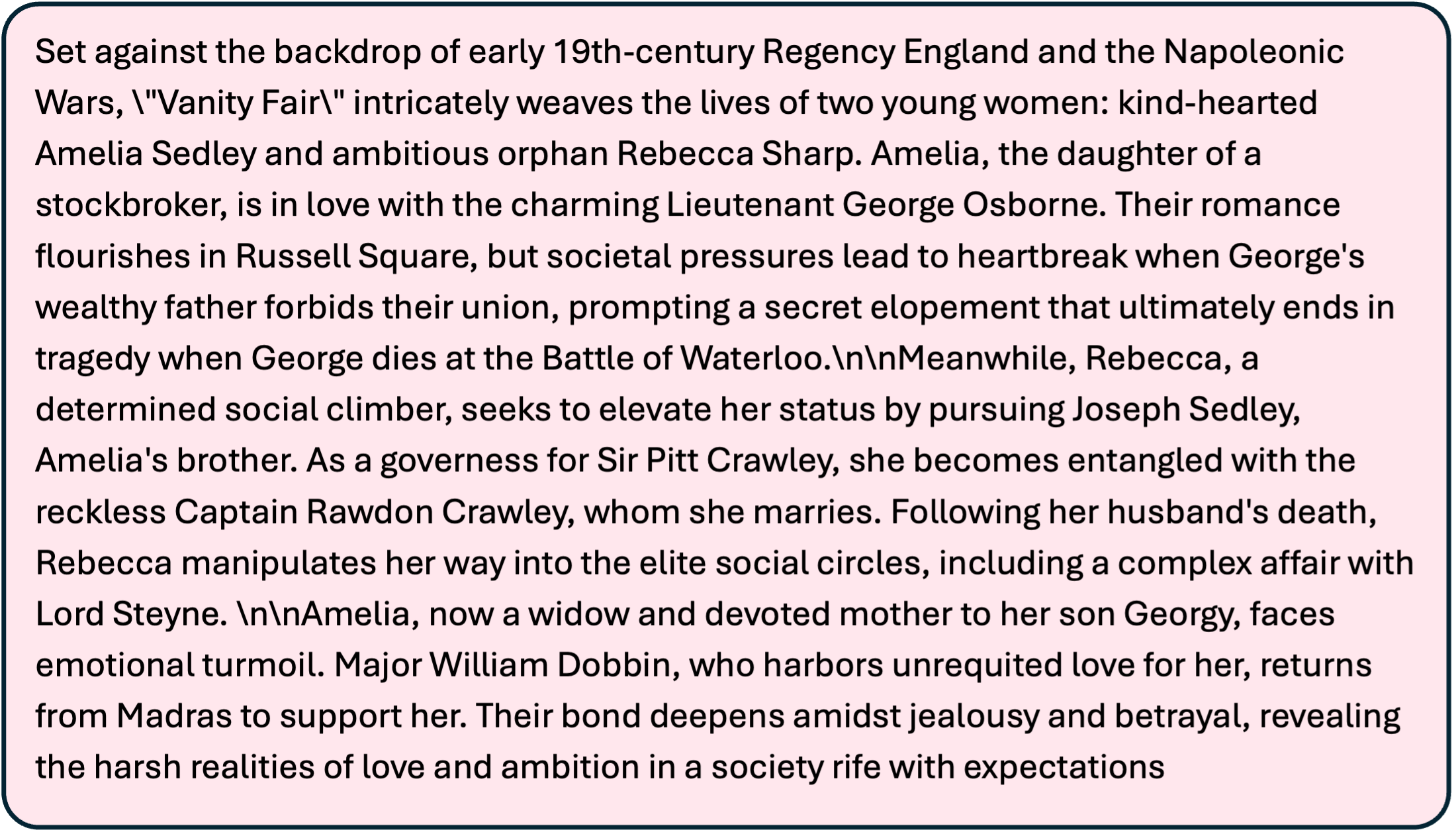}
  \caption{Romance-based summary by GPT + Hier (Vanity Fair).}
  \label{fig:gpt_hier_romance_sum}
\end{figure*}

\begin{figure*}[h!]
  \includegraphics[width=\linewidth]{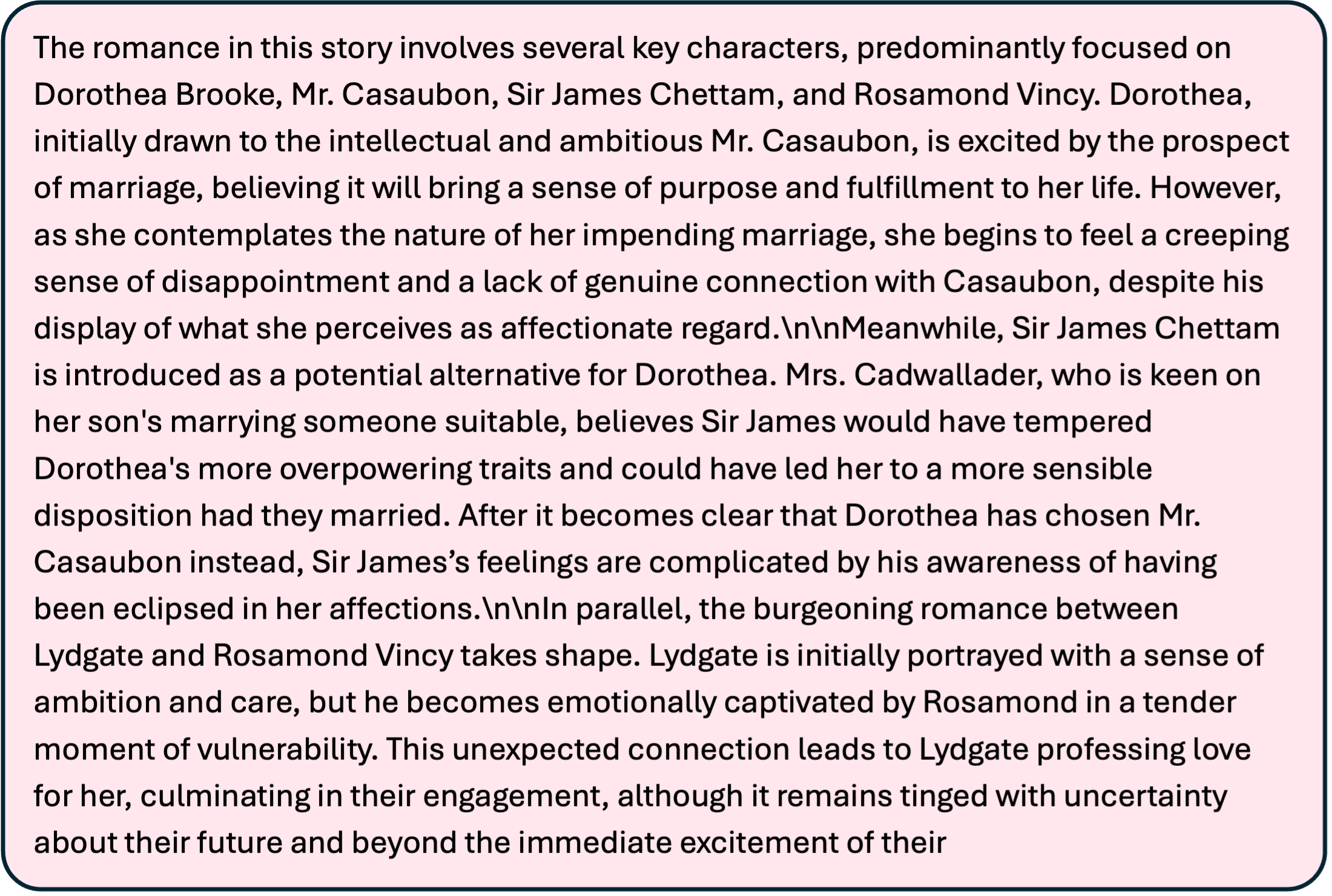}
  \caption{Romance-based summary by NaiveRAG (Vanity Fair).}
  \label{fig:naiverag_romance_sum}
\end{figure*}

\end{document}